\documentclass[times,twocolumn,final]{elsarticle}

\usepackage{arxiv}
\usepackage{graphicx}%
\usepackage{multirow}%
\usepackage{amsmath,amssymb,amsfonts}%

\usepackage{amsthm}%
\usepackage{mathrsfs}%
\usepackage[title]{appendix}%
\usepackage{xcolor}%
\usepackage[utf8]{inputenc}
\usepackage{booktabs}%
\usepackage{algpseudocode,algorithm}
\MakeRobust{\Call}
\usepackage{pifont}
\usepackage{adjustbox}
\usepackage{graphicx}
\usepackage{float}
\usepackage{placeins}
\usepackage{stfloats}
\usepackage{fancyhdr}
\usepackage[hyphens]{url}
\usepackage{hyperref}
\usepackage{caption}

\usepackage{subcaption}
\usepackage{booktabs,tabularx,ragged2e,array,multirow,graphicx}
\newcolumntype{Y}{>{\RaggedRight\arraybackslash}X}
\renewcommand{\arraystretch}{1.15}

\fancypagestyle{firstpagestyle}{
    \fancyhf{} 
    \fancyhead{} 
    \fancyfoot{} 
}

\fancypagestyle{default}{
    \fancyhf{}
    \fancyhead[R]{\thepage} 
}

\begin{document}

\title{Endoshare: A Publicly Available, Surgeons-Friendly Solution to De-Identify and Manage Surgical Videos}

\author[1]{Lorenzo \snm{Arboit}\corref{corresp}}
\cortext[corresp]{Corresponding author: \texttt{larboit@unistra.fr}}
\author[1]{Dennis N. \snm{Schneider}}
\author[1]{Britty \snm{Baby}}
\author[1,2]{Vinkle \snm{Srivastav}}
\author[2,3]{Pietro \snm{Mascagni}\fnref{equal}}
\author[1,2]{Nicolas \snm{Padoy}\fnref{equal}}

\address[1]{University of Strasbourg, CNRS, INSERM, ICube, UMR7357, Strasbourg, France}
\address[2]{IHU Strasbourg, Strasbourg, France}
\address[3]{Fondazione Policlinico Universitario Agostino Gemelli IRCCS, Rome, Italy}

\fntext[equal]{Co-last authors.}

\received{XXX}
\finalform{XXX}
\accepted{XXX}
\availableonline{XXX}
\communicated{XXX}

\begin{abstract}

\textbf{Background:} Video-based assessment and surgical data science can advance surgical training, research, and quality improvement. However, widespread use is limited by heterogeneous recording formats and privacy concerns associated with video sharing. The present work develops, evaluates, and publicly releases \textit{Endoshare}, a surgeon-friendly application that merges, standardizes, and de-identifies endoscopic videos.

\textbf{Methods:} Development followed the software development life cycle with iterative, user-centered feedback. In the analysis phase, an internal survey of clinicians and computer scientists, using 10 usability heuristics, identified early requirements. These findings guided design and implementation toward a cross-platform, privacy-by-design architecture. In the testing phase, an external clinician survey combined the same heuristics with Technology Acceptance Model constructs to evaluate usability and adoption, complemented by a performance assessment across different hardware and configurations. Statistical evaluation used standard descriptive and inferential methods.

\textbf{Results:} In the analysis phase, four clinicians and four computer scientists tested a prototype, reporting high usability (4.68\,{\small$\pm$}\,0.40/5 and 4.03\,{\small$\pm$}\,0.51/5). The lowest score (4.00\,{\small$\pm$}\,0.93/5) concerned the clarity of labels, indicating a need to make functions easier to recognize. The application’s user experience was refined to ensure minimal effort for case selection, video merging, automated out-of-body detection and removal, and filename pseudonymization. In the testing phase, ten surgeons reported high perceived usefulness (5.07\,{\small$\pm$}\,1.75/7), ease of use (5.15\,{\small$\pm$}\,1.71/7), heuristic usability (4.38\,{\small$\pm$}\,0.48/5), and recommendation likelihood (9.20\,{\small$\pm$}\,0.79/10). Benchmarking showed that processing time increased proportionally with video duration and was consistently lower in \textit{fast mode} ($\approx$79\% faster).

\textbf{Conclusions:} \textit{Endoshare} is a publicly available surgeon-friendly solution to manage, standardize, and de-identify surgical videos. \textit{Endoshare} could facilitate the use and sharing of precious data for surgical training, research, and quality improvement. Compliance certification and broader interoperability validation are needed to establish it as a reliable tool for surgical video management.

\end{abstract}

\maketitle
\thispagestyle{firstpagestyle}

\section{Introduction}
\label{sec:introduction}

Minimally invasive procedures, whether laparoscopic or robotic-assisted, have become increasingly integral to surgical practice~\cite{1}, consistently demonstrating significant advantages in postoperative recovery and reduced perioperative morbidity compared to open surgery. 
\\

These minimally invasive procedures are natively guided by endoscopic videos. The analysis of endoscopic videos, often referred to as video-based assessment (VBA), holds substantial value for surgical training~\cite{2,3,4,5}, clinical research~\cite{6,7}, quality improvement~\cite{8,9}, and the standardization of procedural techniques~\cite{10}. 
When VBA is performed across multiple centers, these benefits are amplified by increasing case diversity, enabling inter‐institutional benchmarking, reducing single-site bias, and promoting the development of generalizable competency metrics~\cite{11,12,13,14,15}. 

Despite the recognized importance of multicentric surgical video repositories, substantial barriers hinder the sharing and effective utilization of surgical videos~\cite{16,17}. For instance, the absence of consistent recording protocols results in wide variability in video quality, resolution, frame rates, and file formats~\cite{18}. Such heterogeneity can complicate the development of large-scale, comparable datasets, which are increasingly important for Surgical Data Science (SDS) and surgical Artificial Intelligence (AI) studies. In addition, surgical video recordings are often split into multiple files due to legacy FAT32-based systems, which limit file size to 4 GB and segment recordings as a safeguard against power loss, rendering only the last file vulnerable to corruption. While this ensures recording reliability, it necessitates subsequent file merging and synchronization for efficient full video review and analysis~\cite{18}. 
Finally, surgical videos often include identifiable patient information, such as visible faces, spoken voiceovers, or embedded metadata. Sharing such data requires strict compliance with privacy regulations, including the General Data Protection Regulation (GDPR)~\cite{19} and the Health Insurance Portability and Accountability Act (HIPAA)~\cite{20}. As a result, many institutions are reluctant to share surgical videos unless rigorous de-identification measures are implemented~\cite{21,22,23}. 

Dedicated surgical video platforms addressing some of these issues have emerged~\cite{24}. 
However, adoption remains limited because these systems are typically bundled with costly OR equipment or raise concerns over data custody, vendor lock-in, and contractual access rights~\cite{25}. 
\\

The objective of this study is to develop and evaluate \textit{\textit{Endoshare}}, a novel surgical video management and de-identification application. By releasing this novel solution under a source-available \href{https://polyformproject.org/licenses/noncommercial/1.0.0/}{license}, the study also aims to contribute significantly toward overcoming current barriers and enabling broader adoption of surgical video collection, sharing, and VBA in clinical and research settings.

\section{Materials and Methods}
\label{sec:methods}

This study involved the development and testing of \textit{Endoshare}. The process followed the first four phases of the software development life cycle (SDLC)~\cite{26}: analysis, design, implementation, and testing. Development was iterative and user-centered, incorporating feedback at multiple stages. Two structured surveys were used to guide this process: the first, conducted within the laboratory but with participants not directly involved in the project, served as a usability evaluation tool during the early stages of development; the second involved external surgeons. A technical assessment was conducted to test the performance of \textit{Endoshare} on different computing platforms.

\subsection{Analysis phase}
\label{subsec:analysis}

An initial internal survey, aligned with established SDLC and usability engineering methodologies~\cite{26,27}, was developed during the early development phase to evaluate key system components; this played a central role in the analysis phase to guide further optimization. 
The questionnaire was implemented on Typeform, and the results were analyzed with Microsoft Excel. 
Following Nielsen J.~\cite{28}, it assessed ten heuristic metrics on a 1--5 scale (Visibility, Match, Control, Consistency, Error, Recognition, Flexibility, Aesthetic, Recover, Help; item descriptions in Table~\ref{tab:heuristics_tam_one}), and suggestions were collected for all of them. 
Participants included both clinicians, to assess usability from an end-user perspective, and computer scientists, to identify workflow and technical optimization needs. 
Each participant tested the software prototype and provided feedback through both multiple-choice and open-ended questions.

\begin{table*}[!t]
\centering
\caption{Mapping of original usability and technology acceptance items to study-specific constructs. This table presents each heuristic, perceived usefulness (PU), and perceived ease of use (PEOU) item by its original name, operational definition, and the corresponding remapped label used in the paper.}
\label{tab:heuristics_tam_one}
\setlength{\tabcolsep}{6pt}
\small
\begin{tabularx}{\textwidth}{@{} p{1.1cm} Y Y p{2.6cm} @{}}
\toprule
\textbf{\rotatebox[origin=c]{90}{}} & \textbf{Original Name} & \textbf{Definition} & \textbf{Remapped Name} \\
\midrule
\multirow{10}{*}{\rotatebox[origin=c]{90}{\textbf{Heuristics}}}
& Visibility of System Status & Keep users informed about what's happening through appropriate feedback within a reasonable time frame. & Visibility \\
& Match Between System and Real World & Information appears in a natural and logical order that simplifies the use of the software. & Match \\
& User Control and Freedom & Provide users with clearly marked ``emergency exits'' to undo actions or exit unwanted states easily. & Control \\
& Consistency and Standards & Follow platform conventions and maintain consistency throughout the interface to minimize cognitive load. & Consistency \\
& Error Prevention & Design interfaces to prevent errors by guiding users, offering confirmation dialogs, and providing clear instructions. & Error \\
& Recognition Rather Than Recall & Minimize the need for users to remember information by making objects, actions, and options visible and easily accessible. & Recognition \\
& Flexibility and Efficiency of Use & Accommodate both novice and expert users by offering shortcuts and advanced features without compromising usability for beginners. & Flexibility \\
& Aesthetic and Minimalist Design & Strive for simplicity and clarity in design while avoiding unnecessary elements that could distract or overwhelm users. & Aesthetic \\
& Help Users Recognize, Diagnose, and Recover from Errors & Provide clear error messages and guidance on how to resolve issues when they occur. & Recover \\
& Help and Documentation & Offer comprehensive, easily accessible documentation and support features, while aiming for interfaces intuitive enough to minimize the need for assistance. & Help \\
\midrule
\multirow{6}{*}{\rotatebox[origin=c]{90}{\textbf{PU}}}
& Accomplish Tasks More Quickly & Using the system helps complete tasks faster. & Speed \\
& Improve Job Performance & Using the system enhances job output quality. & Productivity \\
& Increase Productivity & Using the system boosts work throughput. & Performance \\
& Enhance Effectiveness & Using the system makes the user more effective. & Effectiveness \\
& Make it Easier to Do My Job & Using the system simplifies task execution. & Ease \\
& Useful in My Job & The system provides value to job functions. & Usefulness \\
\midrule
\multirow{6}{*}{\rotatebox[origin=c]{90}{\textbf{PEOU}}}
& Easy to Learn & It is easy to learn how to use the system. & Learning Ease \\
& Easy to Control & It is easy to get the system to do what is intended. & Control \\
& Clear and Understandable Interaction & The system is comprehensible in use. & Clarity \\
& Flexible System & The product is flexible to interact with. & Flexibility \\
& Easy to Become Skillful & Skills to use the system can be acquired easily. & Skill \\
& Easy to Use & The system is generally easy to use. & Ease of Use \\
\bottomrule
\end{tabularx}
\end{table*}

\subsection{Design and Implementation phases}
\label{subsec:design_and_implementation}

The application, \textit{Endoshare}, was designed to process surgical videos. Its development was grounded in principles of transparency and long-term adaptability. This strategy ensures that the system can evolve alongside advancements in video processing, AI-based detection of sensitive visual information, and regulatory requirements. Design decisions emphasized flexibility in both deployment and use. The software architecture was developed to accommodate variable hardware capabilities, institutional constraints, and user expertise levels, from researchers with technical backgrounds to clinicians with minimal computing experience. Dependencies were carefully selected to maximize platform compatibility and minimize installation friction, enabling straightforward setup in typical clinical or academic environments.  The project also adopted community-oriented development practices, such as Git-based version control, markdown issue tracking, and standardized code formatting, to maintain clarity and facilitate collaboration. The software is also released under a source-available non-commercial license (PolyForm Noncommercial) to ensure that the codebase remains transparent and reproducible, and can be openly reviewed and extended within the academic community.

\subsection{Testing phase}
\label{subsec:testing}

A second survey was conducted to evaluate the final version of \textit{Endoshare} in external clinical environments, constituting the user-oriented testing phase of the SDLC. Distributed via Google Forms, the survey targeted practicing surgeons who were given early access to \textit{Endoshare}, which was installed on their personal or work computers. On top of the heuristic metrics used in the analysis phase, the design incorporated elements from the Technology Acceptance Model (TAM)~\cite{29} to assess Perceived Usefulness (PU) and Perceived Ease Of Use (PEOU) (Table~\ref{tab:heuristics_tam_one}). The survey included both closed and open-ended questions, grouped into the following domains: user demographics and clinical background, prior experience with video processing, system usability (heuristics), evaluation of specific features (merging and de-identification), adherence to usability principles (TAM), assessment of needs for surgical video management, and overall user satisfaction.
\\

\textit{Endoshare}’s technical performance was evaluated using videos of variable duration and across three different computing environments to assess its generalizability to different file size, compatibility, and processing efficiency under realistic usage conditions. Testing was conducted on a MacBook Pro with an Apple M1 Pro processor, a Windows desktop with an Intel i5 CPU and NVIDIA T1000 GPU, and a high-end Linux workstation equipped with an Intel i7 processor and NVIDIA RTX A5500 GPU. These platforms were selected to reflect a representative range of clinical and research environments: the Intel i5 desktop serves as a conservative baseline for standard hospital workstations, the M1 Pro laptop represents modern high-performance laptops used by senior clinicians, and the i7 workstation models dedicated research or technical laboratory setups. Each system processed three laparoscopic videos of different durations (1, 30, and 60 minutes) across two processing modes (\textit{fast} and \textit{advanced}). Performance was assessed by analyzing processing times.

\begin{figure*}[!t]
\centering
\begin{subfigure}{0.85\textwidth}
  \centering
  \includegraphics[width=\linewidth]{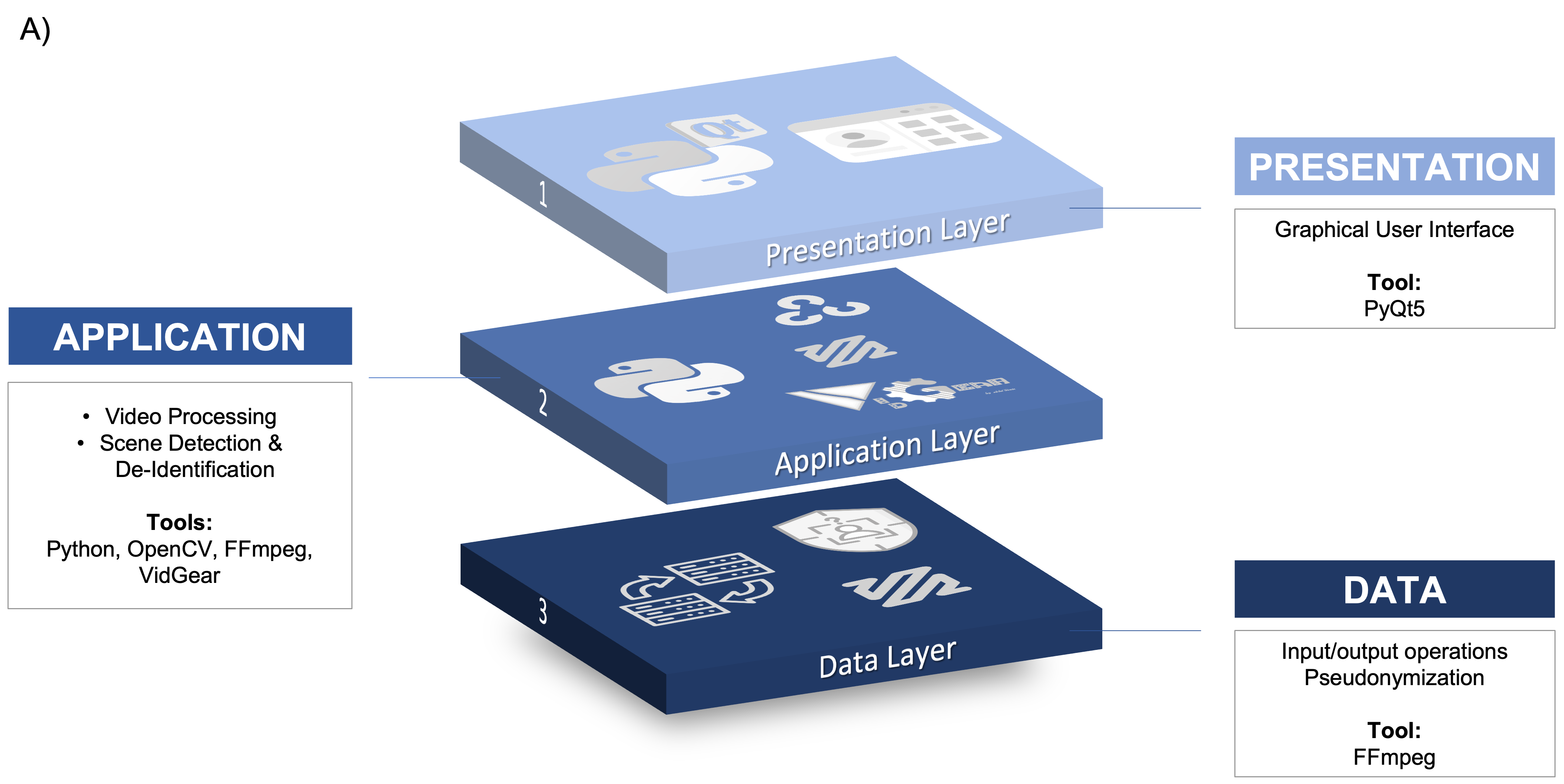}
  \caption{Three‐layered architecture of \textit{Endoshare}, illustrating the Presentation Layer, Application Layer, and Data Layer.}
  \label{fig:architectureA}
\end{subfigure}

\vspace{1em} 
\begin{subfigure}{0.85\textwidth}
  \centering
  \includegraphics[width=\linewidth]{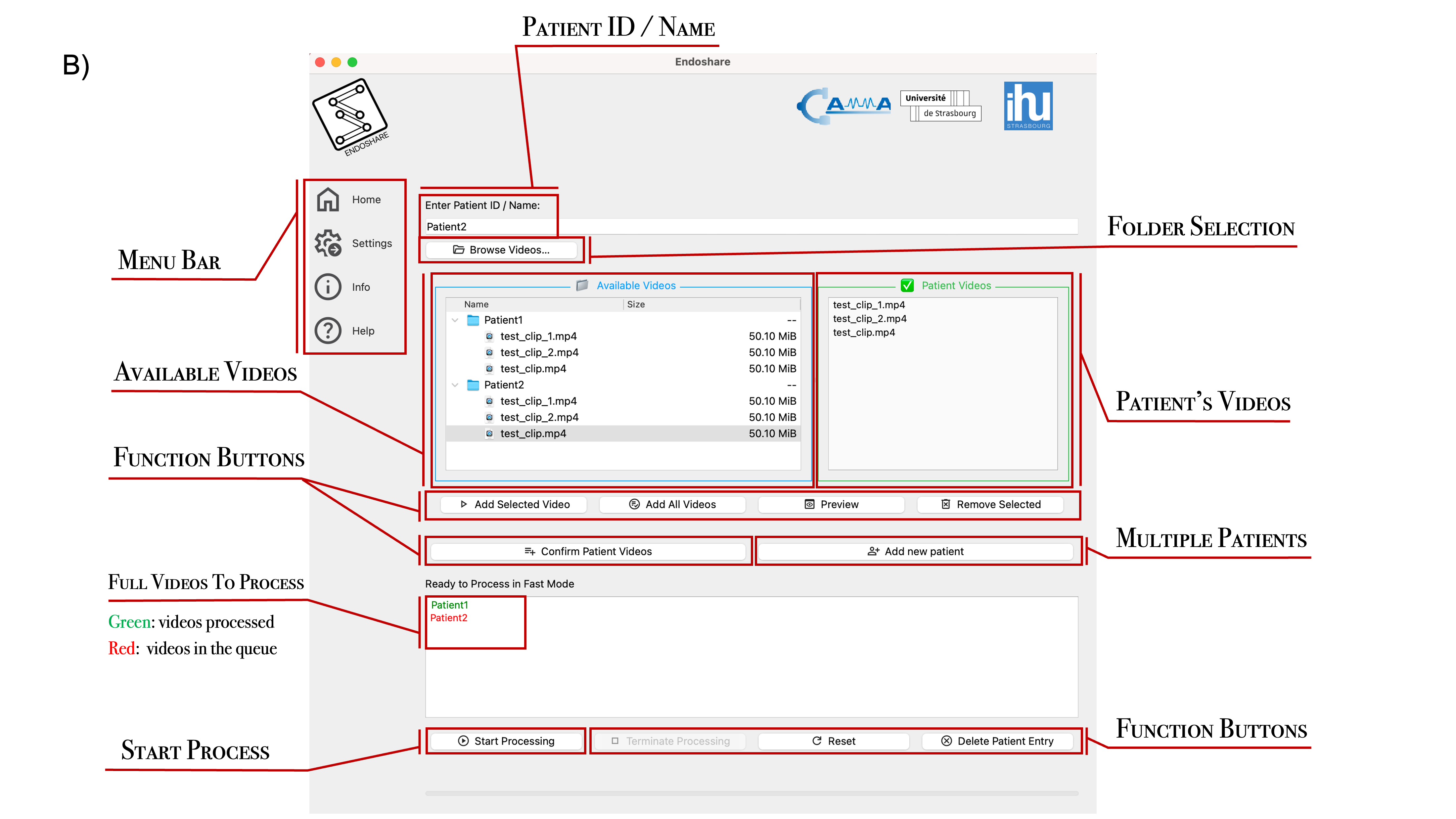}
  \caption{\textit{Endoshare} GUI overview: users enter a patient identifier, select video folders, preview and queue clips for processing, view status indicators for queued files (green = ready, red = in process), and access controls for batch processing.}
  \label{fig:architectureB}
\end{subfigure}

\caption{\textit{Endoshare} architecture and interface. }
\label{fig:1}
\end{figure*}

\begin{figure*}[!t]
  \centering
  \includegraphics[width=\textwidth]{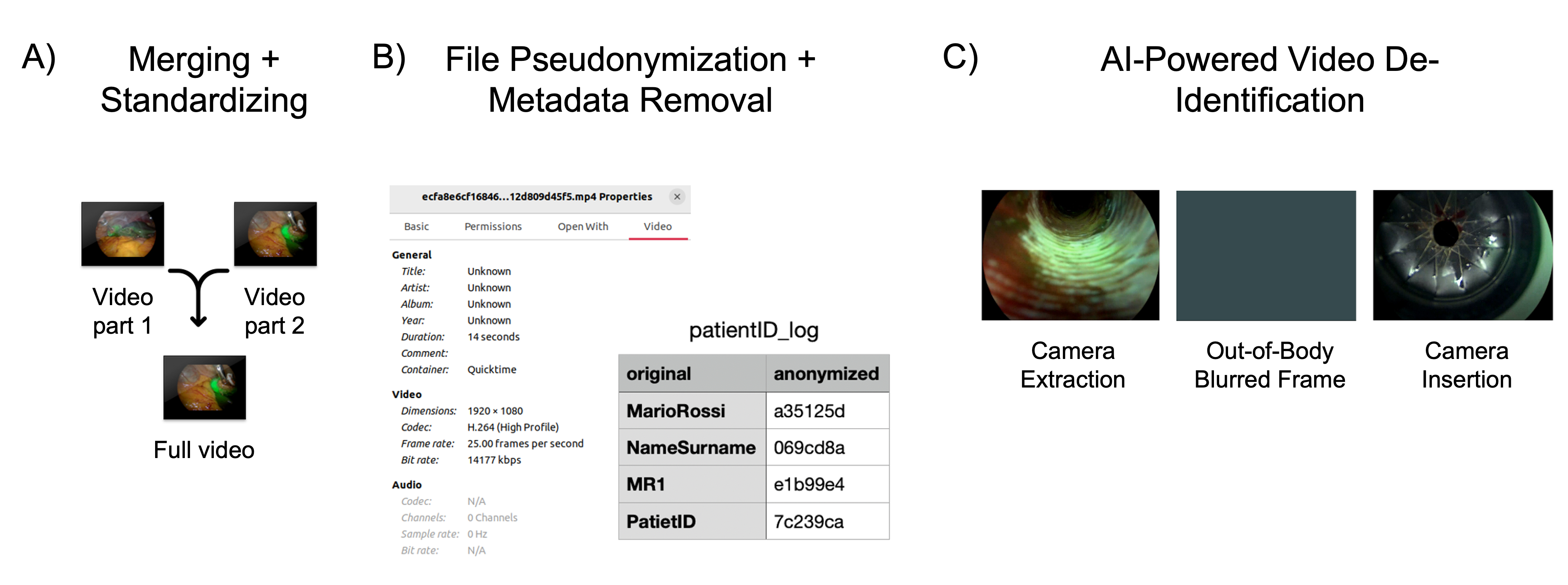}
  \caption{Overview of the \textit{Endoshare} processing pipeline. A) Video merging and standardization combine disparate clips into a single, uniform file; B) File pseudonymization replaces patient identifiers with randomized tokens and strips all embedded metadata; C) AI-powered de-identification detects camera insertion frames, automatically blurs out-of-body scenes, and ensures any remaining identifiable content is masked.}
  \label{fig:2}
\end{figure*}

\subsection{Outcomes and statistical analysis}
\label{subsec:statistical_analysis}

Survey data were analyzed using a mixed-methods approach. Qualitative feedback was subjected to inductive thematic analysis by two independent coders. Quantitative responses were reported as mean ± standard deviation (SD).
For the technical assessment, processing times were measured across three computing platforms (Machine: Mac laptop, Windows desktop, and Linux workstation), two processing configurations (Mode: \textit{fast} and \textit{advanced}), and three video durations (Video: 1, 30, and 60 minutes). Each condition was repeated three times, for a total of 54 benchmark runs. Processing times were summarized for each Machine × Mode × Video condition as geometric mean, which is preferred in this context because it better characterizes variability across multiple runs of the same video on the same machine. Differences between \textit{fast} and \textit{advanced modes} were expressed as geometric mean ratios (\textit{fast / advanced}) with bootstrap 95 \% confidence intervals. To examine scaling behavior, a linear regression through the origin was fitted to model processing time as a function of video duration for each Machine × Mode, yielding slope estimates (seconds per minute) and bootstrap 95 \% CIs. The slope represents the processing rate, that is, how many seconds are required to process one minute of video. All analyses were conducted in Python using \textit{pandas}, \textit{SciPy}, and \textit{Statsmodels}.

\section{Results}
\label{sec:results}

\textit{Endoshare} is a publicly available application, freely downloadable from the following \href{https://camma-public.github.io/Endoshare/}{website}. The source code is available at this \href{https://github.com/CAMMA-public/Endoshare_code/}{GitHub repository}. Thorough documentation is provided to support peer review, reproducibility, and future improvements.

\subsection{Analysis phase}
\label{subsec:analysis}

The first survey was conducted by four surgeons and four computer scientists. Both groups consistently rated the heuristics highly ($4.68/5 \pm 0.40$ vs.\ $4.03/5 \pm 0.51$ for surgeons and computer scientists, respectively). \textit{Error Prevention} was the item that received the highest overall score ($4.63/5 \pm 0.52$), while \textit{Recognition Rather Than Recall} was the lowest ($4.00/5 \pm 0.93$). 

Qualitative analysis of heuristic-based feedback matched the numeric evaluation of these metrics: it revealed positive remarks regarding system consistency, error prevention, and visual minimalism, with users describing the interface as \textit{''easy to follow''} and \textit{''not distracting from work''}. However, comments linked to the heuristics of visibility and recognition highlighted recurring issues with font size, unclear labels, and the lack of visual guidance, suggesting that key elements of the workflow could be made more discoverable and self-explanatory. These items were improved during the development of \textit{Endoshare}.

\subsection{\textit{Endoshare} application}
\label{subsec:application}

The software architecture follows a three-layer design, consisting of a presentation layer, application layer, and data layer (see Figure~\ref{fig:1}A).

The presentation layer provides a graphical user interface (GUI) built with \textit{PyQt5}, enabling access to core functionalities such as video merging, detection and replacement of sensitive scenes, and filename pseudonymization. By encapsulating command-line operations within menu-driven actions and visual prompts, the interface reduces technical complexity and makes the system accessible to clinical end users without programming expertise (Figure~\ref{fig:1}B).

The application layer orchestrates the main processing pipeline using Python and libraries such as \textit{TensorFlow}, \textit{OpenCV}, \textit{FFmpeg}, and \textit{VidGear}. This layer manages all video transformations, including re-encoding, format standardization, and the merging of individual video segments from the same surgical procedure (often split by the laparoscopic tower), as well as the application of deep learning models for automated detection and anonymization of sensitive content. Both processes are illustrated in Figure~\ref{fig:2}A and~\ref{fig:2}C. The pipeline supports both \textit{fast} and \textit{advanced} processing modes, adjusting computational load depending on hardware capabilities and user requirements. \textit{Fast mode} keeps the original video settings and reencodes only the segments with sensitive scenes, making the overall processing much faster, while \textit{advanced mode} enables full control over output specifications by re-encoding the entire video. A core component of the application layer is the integration of a deep learning-based system for out-of-body scene detection, employing a \textit{MobileNetV2} backbone with temporal modeling via \textit{LSTM} layers~\cite{23}. This system classifies video frames to identify those requiring anonymization, supporting automated privacy protection.

The data layer manages input/output operations and ensures the integrity and confidentiality of stored video data. Metadata is automatically stripped from both video containers and streams using \textit{FFmpeg}, and filenames are replaced with pseudonymous identifiers (UUIDs) to mitigate privacy risks (Figure~\ref{fig:2}B). To maintain traceability while preserving privacy, a locally stored spreadsheet maps patient IDs to pseudonyms within the clinic, allowing clinicians to track corresponding cases and selectively share additional anonymized data (such as radiologic or clinical data) with external research laboratories, which never access patient identifiers.

To accommodate diverse user environments, \textit{Endoshare} was developed to run on Windows, macOS, and Linux. Hardware-adaptive processing ensures functionality across devices, from standard laptops to high-performance workstations. 

\textit{Endoshare} also supports multithreaded background processing, enabling batch handling of video datasets while preserving system responsiveness, and allowing prolonged operations, such as overnight processing, to run unattended. Real-time progress indicators and notifications are integrated into the interface to support efficient workflow management. The software is source available and distributed with platform-specific binaries and minimal installation requirements.

\subsection{Testing phase}
\label{subsec:testing}

The testing phase comprised a usability survey and performance benchmarking. The survey received ten responses. The participant population consisted predominantly of male surgical fellows or residents (70\%), primarily in general surgery (90\%), with most respondents working in academic centers (70\%). A majority reported substantial research involvement—60\% leading or co-leading projects—and showed a strong inclination toward technological innovation, with 70\% identifying as early adopters or developers of new technologies (Table~\ref{tab:demographics}).

\begin{table*}[!b]
\centering
\caption{Testing survey responders demographics.}
\label{tab:demographics}
\renewcommand{\arraystretch}{1.15}
\setlength{\tabcolsep}{6pt}
\small
\begin{tabular}{p{4cm} p{6cm} p{4cm}}
\toprule
\textbf{Characteristic} & \textbf{Category} & \textbf{n (\%) or mean~$\pm$~SD} \\
\midrule
Age, years &  & 32.3~$\pm$~4.4 \\
\midrule
\multirow{5}{*}{Country} 
& Italy & 6 (60\%) \\
& Switzerland & 1 (10\%) \\
& Spain & 1 (10\%) \\
& Portugal & 1 (10\%) \\
& United States of America & 1 (10\%) \\
\midrule
\multirow{2}{*}{Gender} 
& Male & 7 (70\%) \\
& Female & 3 (30\%) \\
\midrule
\multirow{2}{*}{Specialty} 
& General Surgery & 9 (90\%) \\
& Obstetrics and Gynecology & 1 (10\%) \\
\midrule
\multirow{2}{*}{Hospital Setting} 
& Academic center & 7 (70\%) \\
& Public hospital & 3 (30\%) \\
\midrule
\multirow{3}{*}{Clinical Experience} 
& Resident & 5 (50\%) \\
& Fellow & 2 (20\%) \\
& Consultant~$\leq$10~y & 3 (30\%) \\
\midrule
\multirow{2}{*}{Academic Qualification} 
& MD & 5 (50\%) \\
& MD + PhD & 5 (50\%) \\
\midrule
\multirow{3}{*}{Scientific Experience} 
& Fellow / PhD Student & 4 (40\%) \\
& Regularly Involved & 1 (10\%) \\
& Lead Projects & 5 (50\%) \\
\midrule
\multirow{2}{*}{Research Years} 
& 1--3 years & 1 (10\%) \\
& 3--10 years & 9 (90\%) \\
\midrule
\multirow{4}{*}{Tech Affinity} 
& Late Adopter & 1 (10\%) \\
& Average Adopter & 2 (20\%) \\
& Early Adopter & 5 (50\%) \\
& Developer & 2 (20\%) \\
\bottomrule
\end{tabular}
\end{table*}

\begin{figure*}[!h]
  \centering
  \includegraphics[width=\textwidth]{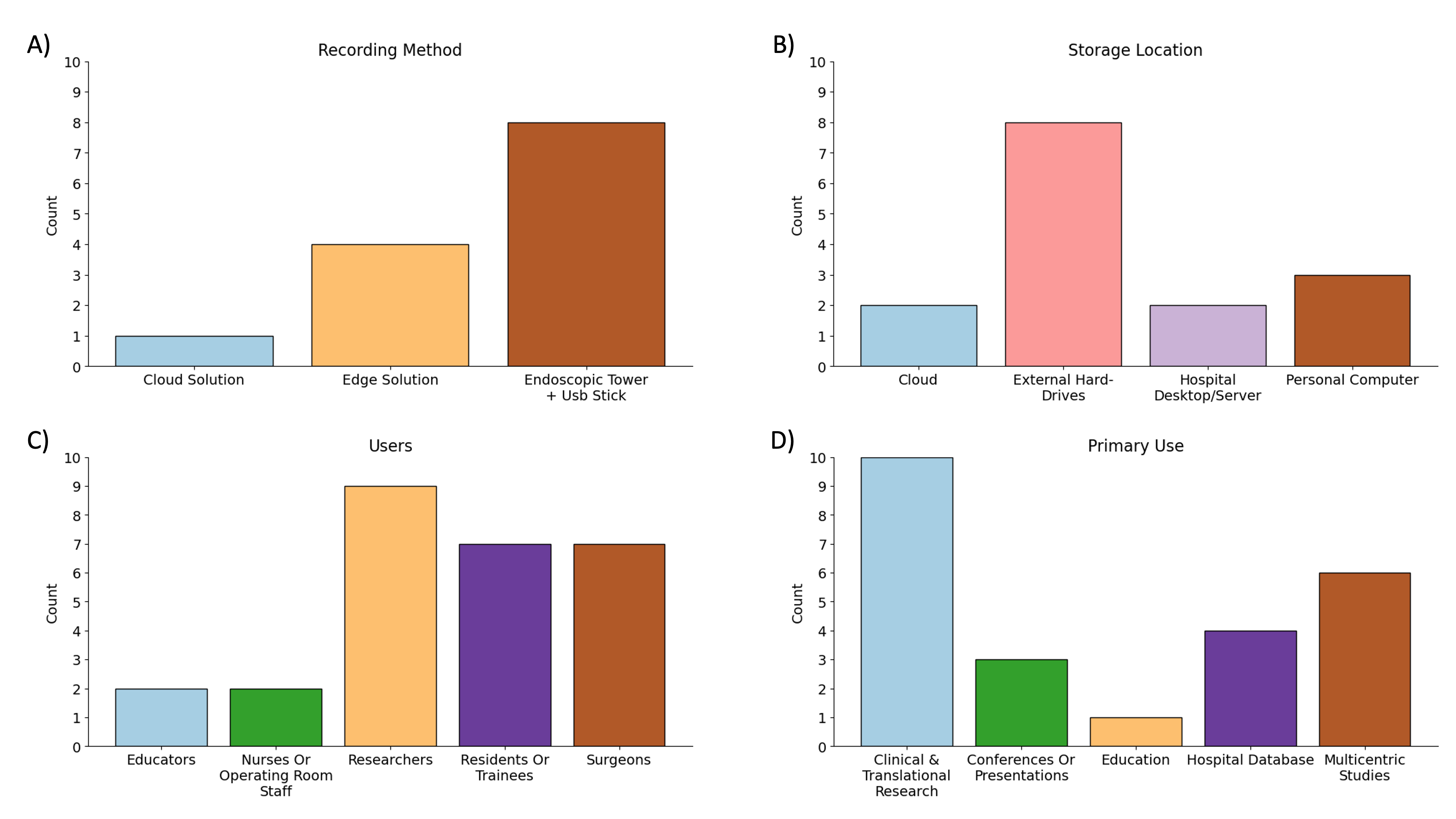}
  \caption{Distribution of user practices in surgical video workflows across four domains. A) preferred recording methods; B) long-term storage locations; C) intended video audiences; and D) reuse contexts for surgical videos.}
  \label{fig:3}
\end{figure*}

All participants recorded surgical videos primarily for research purposes, with secondary uses including conferences (80\%), quality assurance (70\%), training (60\%), documentation (40\%), and education (30\%). The most common capture method was via USB drives connected to the endoscopic tower (80\%), while alternative methods such as edge solutions or cloud upload were used by fewer respondents ($\leq$40\%), as reported in Figure~\ref{fig:3}A. Long-term storage was most frequently on external hard drives (80\%), followed by personal computers (30\%) and hospital servers or cloud platforms (20\%) (Figure~\ref{fig:3}B).

Participants reported intention to use \textit{Endoshare} mainly for clinical and translational research (100\%), followed by multicenter study preparation (60\%), hospital databases (40\%), conferences (30\%), and education (10\%). Those who will mostly use the platform include researchers (90\%), residents or trainees (70\%), and surgeons (70\%), with less involvement from educators (20\%) or nursing/OR staff (20\%) (Figure~\ref{fig:3}C--D).

\begin{figure*}[!h]
  \centering
  \includegraphics[width=\textwidth]{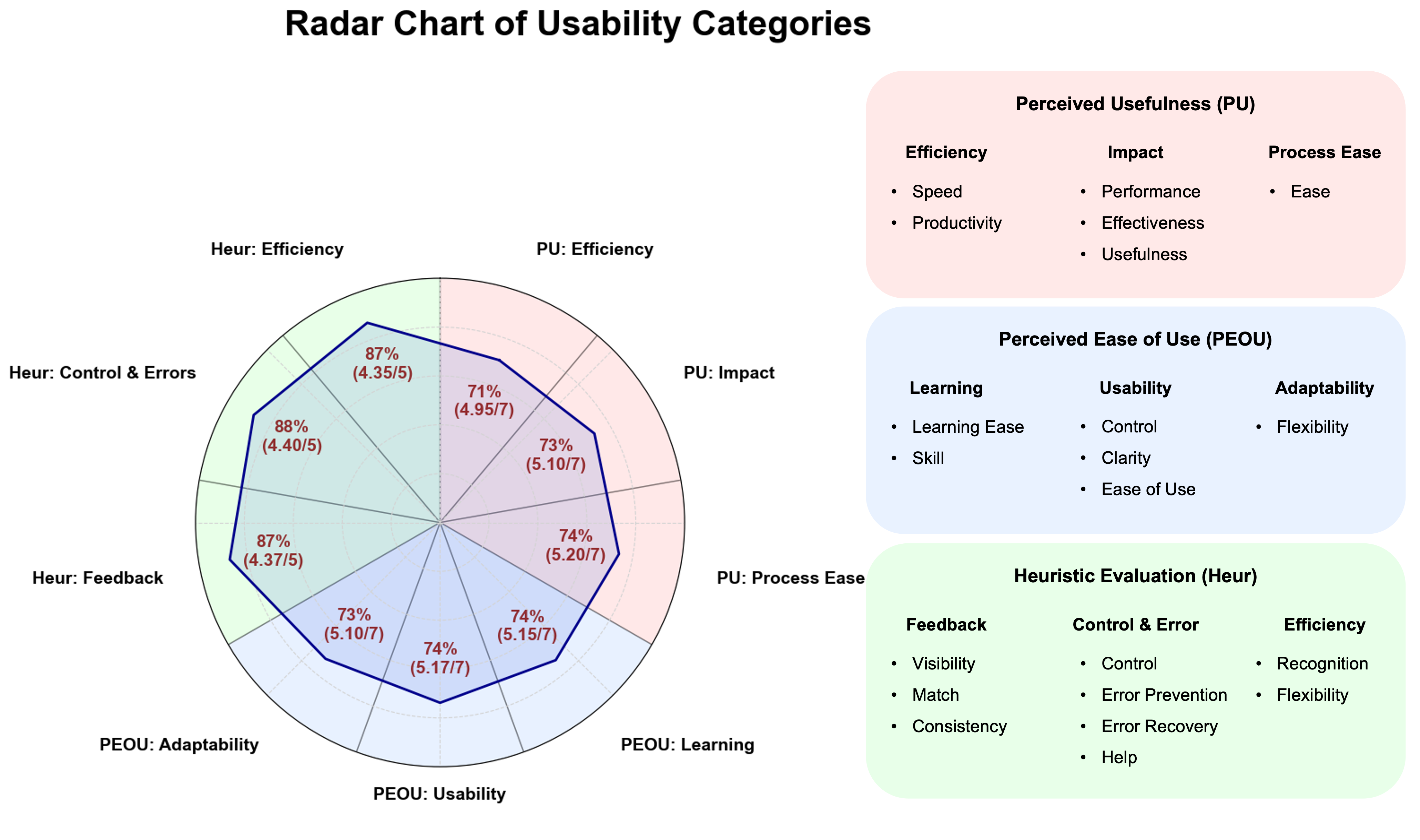}
  \caption{Radar chart summarizing usability evaluations across three domains: Perceived Usefulness (PU), Perceived Ease of Use (PEOU), and Heuristic Usability (Heur). Each axis represents a grouped construct—PU (Efficiency, Impact, Process Ease), PEOU (Learning, Usability, Adaptability), and Heur (Feedback, Control \& Error, Efficiency)—with mean scores plotted (blue line) and annotated values.}
  \label{fig:4}
\end{figure*}

Composite scores showed high system ratings: PU had a mean of $5.07 \pm 1.75$ out of 7, PEOU $5.15 \pm 1.71$ out of 7, heuristic usability $4.38 \pm 0.48$ out of 5, and recommendation likelihood $9.20 \pm 0.79$ out of 10. To enhance visualization, the PU, PEOU, and heuristic evaluation data were aggregated into nine classes (three per item), and the corresponding means for each class are presented in Figure~\ref{fig:4}. All values are reported in the Supplementary Materials (eTable~1).

\begin{figure}[t]
  \centering
  \includegraphics[width=0.5\textwidth]{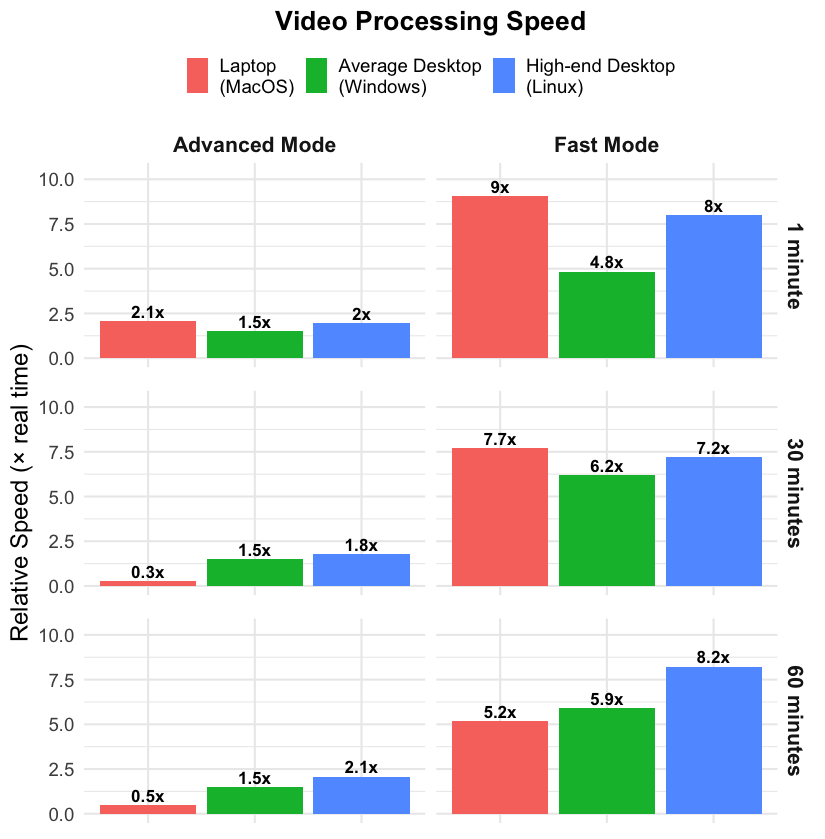}
  \caption{Benchmarking \textit{Endoshare} performance. Processing speed of \textit{Endoshare} across three hardware configurations (macOS laptop, Windows average desktop, Linux high-end desktop) for 1-, 30-, and 60-minute videos in both \textit{fast} and \textit{advanced} modes. Bars indicate geometric mean processing time relative to video duration.}
  \label{fig:5}
\end{figure}

With regard to the performance benchmarking, a total of 18 conditions (3 machines~$\times$~2 modes~$\times$~3 video lengths) were tested, each repeated three times. Geometric mean processing times increased proportionally with video duration on all platforms (Figure~\ref{fig:5}).

Across all machines and video lengths, \textit{fast mode} required only 21.3\% of the processing time of \textit{advanced mode} (pooled geometric mean ratio $=0.213$, 95\%~CI~$0.194$--$0.235$), corresponding to an average 78.7\% reduction in total. 

The linear fits confirmed a near-perfect proportional relationship between processing time and video length. 
Mean slope estimates were $9.22~\mathrm{s/min}$ for \textit{fast mode} and $69.22~\mathrm{s/min}$ for \textit{advanced mode}. 
Processing times varied across hardware, most notably in \textit{advanced mode} for longer videos, while \textit{fast mode} substantially narrowed these gaps, and the relative speed advantage of \textit{fast} over \textit{advanced mode} remained consistent across systems. 
Complete results are reported in Supplementary eTable~2.

\section{Discussion}
\label{sec:discussion}

\begin{figure*}[t]
  \centering
  \includegraphics[width=\textwidth]{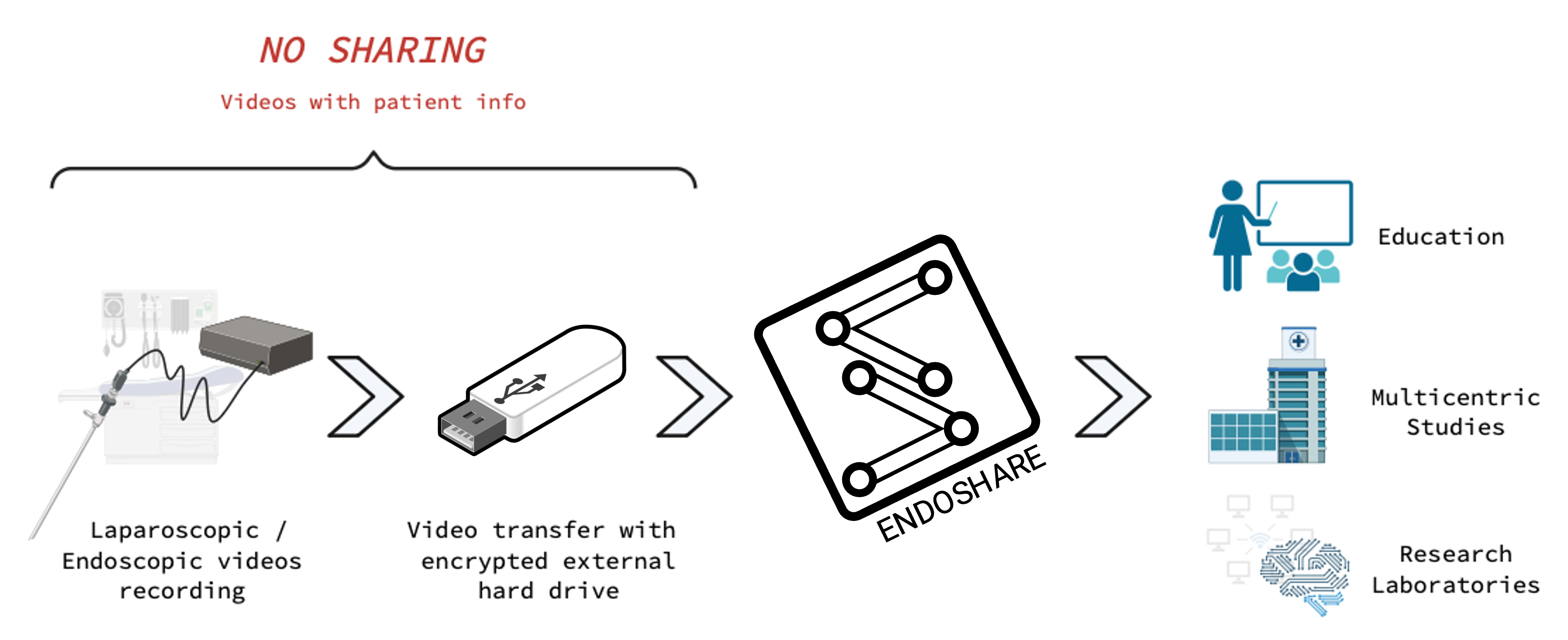}
  \caption{Schematic overview of the \textit{Endoshare} implementation within surgical practice. Endoscopic procedures are recorded and transferred via external storage to any computer with \textit{Endoshare} installed; the platform then automates merging and de-identification, facilitating use and enabling privacy-preserving sharing of videos for education, multicenter research, and laboratory analysis.}
  \label{fig:6}
\end{figure*}

\textit{Endoshare}, a publicly available and platform-independent application, systematically addresses main limitations in surgical video sharing and management. Developed through an iterative, user-centered design approach, \textit{Endoshare} is an easy to use, surgeon-friendly solution to streamline surgical video use, as illustrated in Figure~\ref{fig:6}.

The pipeline starts with the recording of videos guiding minimally invasive procedures; users can then transfer via encrypted external hard drives the videos to any computer equipped with \textit{Endoshare}. With a single setup step, all videos are automatically prepared for processing and export, making them immediately available for educational or research use. By default, the application launches in \textit{fast mode} to maximize throughput, while an advanced interface offers power users fine‐grained control over encoding parameters, such as resolution and frame rate, when needed. A step‐by‐step user guide is bundled with the software to facilitate initial installation and setup.

This application was shaped and improved through two distinct evaluation phases. In the analysis phase, the initial usability assessment guided early refinements of \textit{Endoshare}, particularly in visibility, recognition, and labeling. In the subsequent testing phase, external users confirmed the strengths of the application, especially in error prevention, interface consistency, and minimalistic design, which were reflected in the high heuristic usability scores. In addition, high PU and PEOU scores, together with strong recommendation likelihood, confirmed high user acceptance. Complementing these usability findings, the technical evaluation confirmed \textit{Endoshare}’s consistent performance across different hardware configurations and video durations. Processing time scaled proportionally with video length, and the system maintained stable operation on all tested machines, demonstrating robustness and cross-platform compatibility.

\textit{Endoshare} is characterized by an open architecture, minimal hardware dependencies, and ease of deployment. To the best of our knowledge, it represents the first freely available application in this domain, whereas comparable systems are generally integrated within proprietary commercial packages or distributed at cost. \textit{Endoshare} can be seamlessly installed on any Windows, macOS, or Linux workstation and operates using standard file-system protocols. Its modular design permits rapid inclusion of new codecs or de-identification algorithms without altering core infrastructure, and its reliance on widely used libraries (\textit{PyQt5}, \textit{FFmpeg}, \textit{TensorFlow}) ensures transparency and reproducibility. In contrast, established clinical video systems often impose licensing fees, rigid workflows, and limited customization, which can hinder adoption, particularly in resource-constrained or research-focused settings. By providing a fully auditable codebase and comprehensive user documentation, \textit{Endoshare} lowers the barrier for centers to implement video sharing while keeping data management fully within the institution.

\subsection{Future Developments}
\label{subsec:future_developments}

Several developments are planned to enhance the functionality, usability, and robustness of \textit{Endoshare}. Efforts are underway to ensure that the application can dynamically leverage available GPU resources across heterogeneous computing environments. This includes configuring deep learning inference and video processing pipelines to operate efficiently regardless of the host system’s hardware configuration.

Functionality will also be extended to include manual review and correction of frames misclassified by the automated de-identification algorithm. A dedicated interface will allow users to inspect flagged scenes, adjust classifications, and reinitiate the de-identification process as needed. This manual override mechanism is critical for addressing edge cases and improving the overall reliability and trustworthiness of the platform in clinical settings.

\subsection{Limitations}
\label{subsec:limitations}

Despite these promising findings, this study presents several limitations. The modest sample sizes of the usability surveys limit the statistical power and generalizability of conclusions drawn from user feedback. Future studies with larger, more diverse cohorts are essential to robustly validate these preliminary results and to better understand subgroup-specific usability needs. Additionally, the technical assessments conducted on three computing environments, although comprehensive, may not fully represent the broader spectrum of machines. To address both of these limitations, we will release \textit{Endoshare}’s full codebase and provide a Google Form for user feedback. Through this form, users can share their experience and upload an automatically generated log file capturing processing speed, hardware specifications, and any errors encountered. No personal identifiers will be collected, and all data will be used exclusively for research to guide the development of a more stable, refined version of \textit{Endoshare}. Please note that submissions are for research purposes only: individual support or troubleshooting based on these data will not be provided.

Furthermore, the current AI model is limited to detecting and removing out‐of‐body scenes and does not yet identify or obscure sensitive on‐screen elements such as patient names, dates, or institutional information occasionally embedded by endoscopic towers.

Finally, regional variations in regulatory frameworks complicate the establishment of standardized international video‐sharing practices~\cite{30,31,32}. Although \textit{Endoshare} streamlines preprocessing and de‐identification workflows, it cannot, by itself, guarantee compliance across all jurisdictions.

\section{Conclusions}
\label{sec:conclusions}

Overall, \textit{Endoshare} demonstrates significant potential as a standardized and efficient surgical video‐sharing tool by addressing both technical and organizational dimensions, from rigorous de‐identification to user‐informed design, cross‐platform optimization, and scalable architecture. Continued community engagement, larger‐scale validations, and strategic interoperability initiatives will be essential to realize its full potential as a foundational tool in surgical education, research, and quality improvement.

\section{Disclosures}

Pietro Mascagni and Nicolas Padoy are co-founders and shareholders of Scialytics. All other authors have no conflicts of interest to report.

\section{Acknowledgement}

We thank the respondents to our two surveys, listed in alphabetical order among those who provided consent: Andrea Balla, Francesco Brucchi, Keqi Chen, Laleh Foroutani, Alain Garcia, Marta Goglia, Pooja Jain, Joël L. Lavanchy, Giuseppe Massimiani, Adrien Meyer, Aditya Murali, Matteo Pavone, Saurav Sharma, Antonio Sampaio Soares, Maria Vannucci, Alice Zampolini.

\subsection{Data Availability Statement}

The application is freely available for download from the official \href{https://camma-public.github.io/Endoshare/}{website}, with its complete source code openly accessible on \href{https://github.com/CAMMA-public/Endoshare_code/}{GitHub}. The following \href{https://forms.gle/ndkDNeNzSUfSYmPT9}{Google Form} can be used for user feedback.

\subsection{Funding Statement}

This work was developed within the Interdisciplinary Thematic Institute HealthTech (ITI 2021-2028 program of the University of Strasbourg, CNRS and Inserm),  supported by IdEx Unistra (ANR-10-IDEX-0002) and SFRI (STRAT’US project, ANR-20-SFRI-0012) under the framework of the French Investments for the Future Program.

This work has received funding from the European Union (ERC, CompSURG, 101088553). Views and opinions expressed are however those of the authors only and do not necessarily reflect those of the European Union or the European Research Council. Neither the European Union nor the granting authority can be held responsible for them. This work was also partially supported by French state funds managed by the ANR under Grant ANR-10-IAHU-02.

\subsection{CRediT Authorship Contribution Statement}

LA: \textit{Conceptualization, Methodology, Software, Validation, Formal analysis, Investigation, Data curation, Writing - Original Draft, Writing - Review \& Editing, Visualization.} DS, BB, VS: \textit{Methodology, Software, Writing - Review \& Editing.} PM: \textit{Conceptualization, Methodology, Writing - Review \& Editing, Supervision.} NP: \textit{Conceptualization, Writing - Review \& Editing, Supervision, Resources, Funding Acquisition, Project Administration.}

\bibliographystyle{splncs04}
\bibliography{arxiv}

\newpage
\setcounter{table}{0}
\renewcommand{\thetable}{\arabic{table}}
\captionsetup[table]{labelformat=simple, labelsep=period, name={eTable}}

\newcommand{\CI}[2]{[#1--#2]} 
\newcommand{\minu}{\,\mathrm{min}} 
\newcommand{\spermin}{\,\mathrm{s/min}} 

\begin{table*}[t]
\centering
{\LARGE \bfseries Supplementary Materials}\par\vspace{3em}
\caption{Mean and standard deviation (SD) of usability evaluation scores across three domains: heuristic usability (Heuristics, out of 5), perceived usefulness (PU, out of 7), and perceived ease of use (PEOU, out of 7).}
\label{etab:1}
\setlength{\tabcolsep}{4pt}\small
\begin{tabularx}{0.6\linewidth}{@{} p{0.9cm} Y p{2.0cm} @{}}
\toprule
\textbf{\rotatebox[origin=c]{90}{Domain}} & \textbf{Variable} & \textbf{Mean $\pm$ SD} \\
\midrule
\multirow{9}{*}{\rotatebox[origin=c]{90}{\textbf{Heuristics (out of 5)}}}
& Visibility          & 4.3~$\pm$~0.8 \\
& Match               & 4.5~$\pm$~0.7 \\
& Control             & 4.5~$\pm$~0.5 \\
& Consistency         & 4.3~$\pm$~0.5 \\
& Error Prevention    & 4.5~$\pm$~0.7 \\
& Recognition         & 4.3~$\pm$~0.7 \\
& Flexibility         & 4.4~$\pm$~0.7 \\
& Error Recovery      & 4.2~$\pm$~0.9 \\
& Help                & 4.4~$\pm$~0.7 \\
\midrule
\multirow{6}{*}{\rotatebox[origin=c]{90}{\textbf{PU (out of 7)}}}
& Speed               & 5.0~$\pm$~1.8 \\
& Performance         & 4.9~$\pm$~1.8 \\
& Productivity        & 4.9~$\pm$~1.8 \\
& Effectiveness       & 5.1~$\pm$~1.7 \\
& Ease                & 5.2~$\pm$~1.8 \\
& Usefulness          & 5.3~$\pm$~1.8 \\
\midrule
\multirow{6}{*}{\rotatebox[origin=c]{90}{\textbf{PEOU (out of 7)}}}
& Learning Ease       & 5.2~$\pm$~1.7 \\
& Control             & 5.1~$\pm$~1.7 \\
& Clarity             & 5.2~$\pm$~1.8 \\
& Flexibility         & 5.1~$\pm$~1.7 \\
& Skill               & 5.1~$\pm$~1.7 \\
& Ease of Use         & 5.2~$\pm$~2.0 \\
\bottomrule
\end{tabularx}
\end{table*}

\begin{table*}[t]
\centering
\caption{\textbf{Performance evaluation of \textit{Endoshare}.} 
Panel~(A) reports descriptive statistics for processing times, including geometric means (seconds) and slope estimates (seconds per minute, 95\%~CI) obtained from linear models through the origin fitted separately for each Machine~$\times$~Mode. 
Panel~(B) summarizes relative performance between \textit{fast} and \textit{advanced} modes using geometric mean ratios (GMR, fast / advanced) with bootstrap 95\%~confidence intervals and slope ratios. 
Consistent slope and GMR values across video lengths indicate proportional (linear) scaling of processing time with video duration and a stable relative speed advantage of \textit{fast} mode across hardware configurations.}
\label{etab:2}
\small
\setlength{\tabcolsep}{6pt}
\renewcommand{\arraystretch}{1.15}
\begin{tabular}{@{} l l l c c @{}}
\toprule
\multicolumn{3}{c}{\textbf{Machine / Mode / Video}} &
\multicolumn{1}{c}{\textbf{Geometric Mean (s)}} &
\multicolumn{1}{c}{\textbf{Slope (s/min) [95\% CI]}} \\
\midrule

\multirow{6}{*}{\textbf{Average Desktop (Windows)}}
 & \multirow{3}{*}{Advanced}
  & 1\,min  & 39.4     & 40.25 (39.63--40.86) \\
 & & 30\,min & 1202.7   & --- \\
 & & 60\,min & 2416.8   & --- \\
 & \multirow{3}{*}{Fast}
  & 1\,min  & 12.4     & 10.04 (9.96--10.13) \\
 & & 30\,min & 290.8    & --- \\
 & & 60\,min & 607.3    & --- \\
\midrule

\multirow{6}{*}{\textbf{High-end Desktop (Linux)}}
 & \multirow{3}{*}{Advanced}
  & 1\,min  & 30.2     & 29.82 (28.96--30.65) \\
 & & 30\,min & 1010.2   & --- \\
 & & 60\,min & 1731.0   & --- \\
 & \multirow{3}{*}{Fast}
  & 1\,min  & 7.5      & 7.48 (6.77--8.18) \\
 & & 30\,min & 249.6    & --- \\
 & & 60\,min & 435.3    & --- \\
\midrule

\multirow{6}{*}{\textbf{Laptop (macOS)}}
 & \multirow{3}{*}{Advanced}
  & 1\,min  & 29.1     & 137.60 (101.98--190.23) \\
 & & 30\,min & 5729.8   & --- \\
 & & 60\,min & 7213.7   & --- \\
 & \multirow{3}{*}{Fast}
  & 1\,min  & 6.6      & 10.14 (5.93--16.02) \\
 & & 30\,min & 227.3    & --- \\
 & & 60\,min & 619.2    & --- \\
\midrule

\multicolumn{5}{c}{\textbf{(A) Fast vs Advanced}} \\
\midrule
\textbf{Machine} & \multicolumn{2}{l}{\textbf{GMR (Fast/Advanced) [95\% CI]}} &
\multicolumn{2}{l}{\textbf{Slope ratio (Fast/Advanced)}} \\
\midrule
Average Desktop (Windows) & \multicolumn{2}{l}{0.269 (0.265--0.274)} & \multicolumn{2}{l}{0.249} \\
High-end Desktop (Linux)  & \multicolumn{2}{l}{0.250 (0.225--0.274)} & \multicolumn{2}{l}{0.251} \\
Laptop (macOS)            & \multicolumn{2}{l}{0.120 (0.092--0.158)} & \multicolumn{2}{l}{0.074} \\
\bottomrule
\end{tabular}
\end{table*}

\end{document}